\documentclass[lettersize,journal]{IEEEtran}
\usepackage{amsmath,amsfonts}
\usepackage{algorithmic}
\usepackage{algorithm}
\usepackage{array}
\usepackage[caption=false,font=normalsize,labelfont=sf,textfont=sf]{subfig}
\usepackage{textcomp}
\usepackage{stfloats}
\usepackage{url}
\usepackage{verbatim}
\usepackage{graphicx}
\usepackage{booktabs}
\usepackage{cite}
\usepackage{graphicx} 
\usepackage{xcolor}   
\usepackage{makecell} 
\usepackage{multirow}
\usepackage{xcolor}

\begin{document}

\title{IAE-VTG: Interaction-Aligned Action–Entity Video Temporal Grounding}

\author{Shiwen Zhao$^{1}$,  Qi Zhang $^{2,*}$,~\IEEEmembership{Graduate Student Member,~IEEE}, Sezer Karaoglu$^{2}$, Theo Gevers$^{2}$, Martin R. Oswald$^{2}$

\thanks{$^{1}$Shiwen Zhao is with School of Computer Science, The University of Sydney, Sydney, NSW 2006, Australia. (e-mail: szha0669@uni.sydney.edu.au).}%

\thanks{$^{2}$Qi Zhang, Sezer Karaoglu, Theo Gevers and  Martin R. Oswald are with the Informatics Institute at University
of Amsterdam, 1098 XH Amsterdam, the Netherlands. (e-mail: q.zhang2@uva.nl,s.karaoglu@uva.nl, th.gevers@uva.nl, m.r.oswald@uva.nl).}%

\thanks{$^{*}$Project Leader And Corresponding author: Qi Zhang (e-mail: q.zhang2@uva.nl).}
}


\markboth{IEEE Transactions on Image Processing}%
{Zhao \MakeLowercase{\textit{et al.}}: Interaction-Aligned Video Temporal Grounding}


\maketitle

\begin{abstract}
Video Temporal Grounding (VTG) localizes the video segment that matches a natural-language query. Many queries describe an action performed by a particular entity. Existing methods often encode the query as a whole or use general video-text interactions, without explicitly checking whether the action and entity occur together. They may therefore select a segment that contains both concepts but not the event described by the query. We propose Interaction Aligned Action-Entity Video Temporal Grounding (IAE-VTG), which models this relationship at both the representation and training assignment levels. First, the Fine-grained Disentangled Interaction Module (FDIM) separates action and entity related query information and aligns it with complementary motion and appearance features. It then combines token-level interactions to build representations that capture the relationship between the action and entity. Second, Interaction-Sensitive Assignment (ISA) adds this interaction evidence to bipartite matching, so training targets are selected using both temporal overlap and semantic compatibility. This reduces supervision from temporally plausible but semantically incorrect proposals. Experiments on QVHighlights, Charades-STA, and TACoS show that IAE-VTG consistently improves strong baselines and achieves competitive or state-of-the-art performance on standard grounding metrics. Additional analyses show that the method is especially effective when similar actions or entities appear at multiple times and produces more reliable assignments for complex events.

\end{abstract}


\section{Introduction}
Video Temporal Grounding (VTG)~\cite{Alpher03, Alpher79, Alpher05, Alpher06, Alpher07, Alpher08, Alpher09, Alpher24, Alpher31} aims to localize the temporal segment in a video that corresponds to a natural language query. 
By establishing fine grained correspondence between visual content and queries, VTG serves as an important component of multimodal video understanding and has attracted increasing attention.

A grounding prediction can be semantically correct while remaining temporally incorrect. For example, a model may recognize the queried entity and respond to all frames in which it appears, even though the queried action occurs only during a short subinterval.
The resulting prediction contains relevant visual concepts but fails to recognize when they form the queried action.
This reveals an important distinction between \emph{concept relevance} and \emph{interaction consistency}.  \emph{Interaction consistency} means recognizing an action or entity independently and does not establish that they are compositionally associated within the localized moment.

\begin{figure}[t]
    \centering
    \includegraphics[width=.95\linewidth]{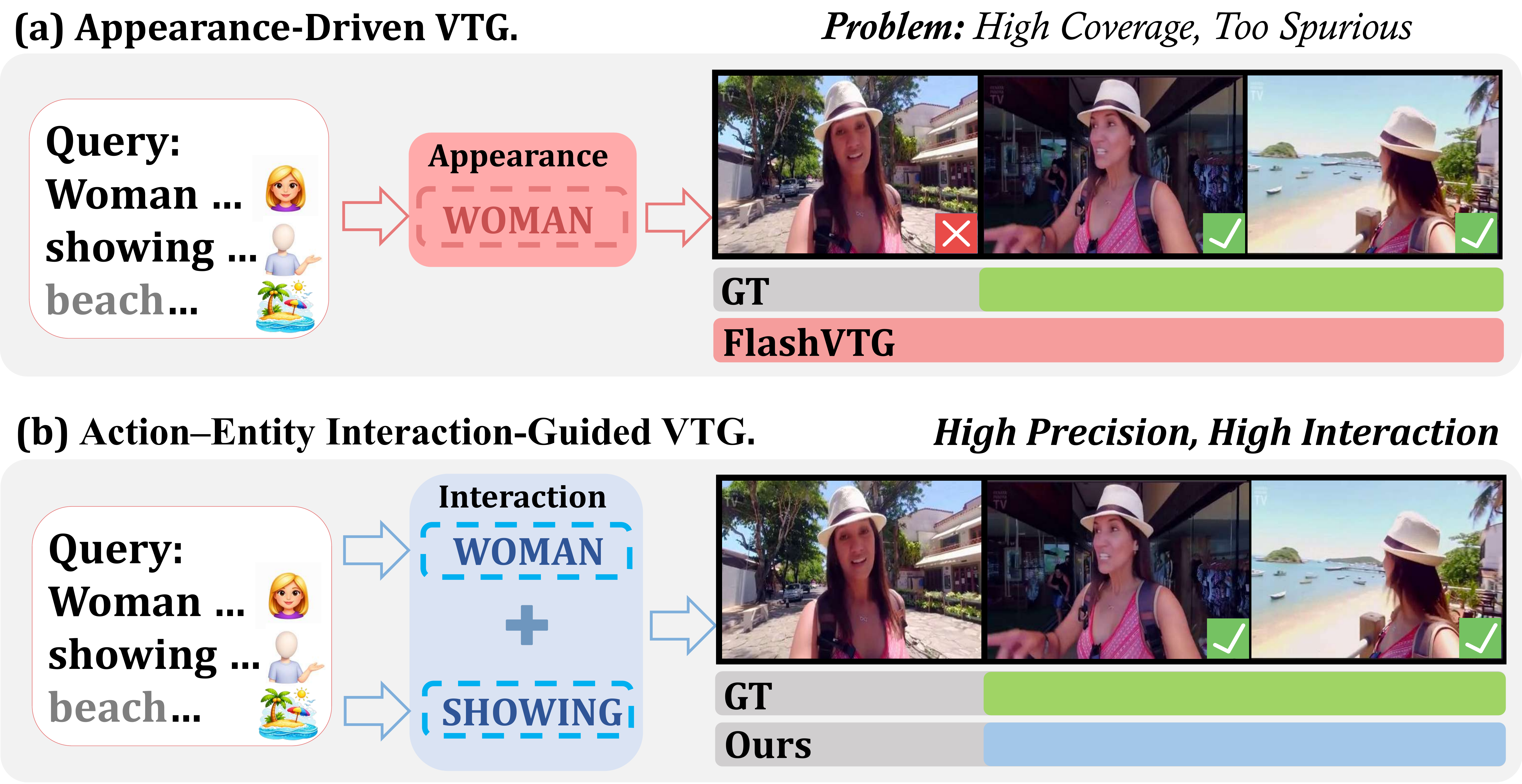}
    \caption{
    Comparison between appearance dominant grounding and the proposed interaction aware grounding.
    FlashVTG~\cite{Alpher31} responds broadly to the salient entity, whereas IAE-VTG jointly considers appearance and motion evidence to localize the queried action-entity interaction.
    Colored curves denote predicted grounding responses, and the green interval indicates the ground-truth moment.
    }
    \label{fig:motivation}
\end{figure}

Existing VTG methods can be divided into proposal based and proposal free paradigms~\cite{Alpher10, Alpher11, Alpher12}.
Proposal based methods~\cite{Alpher06, Alpher07, Alpher08, Alpher09} generate predefined or learned candidate segments and rank them according to video and text relevance. The proposal free methods predict temporal boundaries without densely enumerating temporal candidates.
More recently, DETR-based~\cite{Alpher79, Alpher34} approaches formulate VTG as a set
prediction problem. It means learnable moment queries decode a set
of candidate temporal moments and bipartite matching determines
their supervision.
Despite their architectural differences, these methods identify target moments through query relevant visual evidence. But they would fail When actions recur at different temporal locations and occupie only a small fraction of the entity visible interval.
Appearance cues identify the correct entity while providing limited temporal information. The motion cues indicate a relevant action without determining the entity involved.
Many recent VTG approaches~\cite{Alpher79, Alpher31, Alpher34} employ token level or fine grained cross modal interactions and can capture both action related and entity related evidence.
Recent works have also explored phrase level~\cite{Alpher32} or event level grounding~\cite{Alpher38}.
However, these cues are still commonly integrated through correlation driven matching, without explicitly verifying whether the queried action and entity are jointly supported within the same temporal context.
Consequently, a segment may receive a high grounding score because it contains a salient entity or a similar action, even though it does not contain the both.


Figure~\ref{fig:motivation} provides a concrete example of this failure mode.
Given the query \textit{``A woman in a pink dress and white hat showing off views of the beach she is at,''} FlashVTG~\cite{Alpher31} responds strongly to the persistent entity cue, \emph{woman}, and consequently produces an overly broad temporal prediction.
However, the subject remains visible outside the interval in which the queried action, \emph{showing}, is performed.
Accurate localization therefore requires more than detecting relevant entities or actions independently.
It requires determining whether they form the queried interaction at each temporal location.

To address the problem, we propose IAE VTG, an interaction aware framework for fine grained temporal grounding.
IAE VTG exploits complementary appearance and motion streams, which provide different inductive biases for entity related and action related evidence.
We introduce a Fine grained Disentangled Interaction Module (FDIM) that grounds action related and entity related query tokens in these complementary streams and explicitly models their cross stream interactions.
Rather than treating the two types of evidence as independent sources of relevance, FDIM constructs composition sensitive representations that indicate whether they are jointly supported within the same temporal context.

We further develop an Interaction Sensitive Assignment (ISA) strategy that incorporates interaction consistency into bipartite matching.
ISA augments the matching cost with a consistency term derived from binding and saliency evidence, discouraging predictions with high localization confidence but incomplete action and entity support from receiving positive supervision.
In this way, FDIM improves the semantic structure of the representations, while ISA ensures that the training assignments reflect the same interaction requirement.

The main contributions of this work are summarized as follows:
\begin{itemize}
\item We identify action-entity ambiguity as an important source of spurious temporal grounding and propose IAE-VTG, an interaction-aware framework that explicitly models whether the queried action and entity jointly occur within a candidate moment. We use various datasets~\cite{Alpher79, Alpher14, Alpher42} to demonstrate that IAE-VTG achieves state-of-the-art or competitive performance across standard grounding metrics.

\item We introduce FDIM, which grounds action and entity related query tokens in complementary motion and appearance streams and models their fine-grained interactions to construct composition-sensitive video--text representations.

\item We develop ISA, an interaction sensitive assignment strategy that incorporates binding--saliency consistency into bipartite matching, thereby producing semantically more reliable training assignments.
\end{itemize}

\section{Related Work}

\subsection{Video Temporal Grounding}
Video Temporal Grounding~\cite{Alpher03,Alpher79,Alpher05,Alpher06,Alpher07,Alpher08,Alpher09,Alpher13, Alpher14, Alpher15,Alpher16, Alpher17, Alpher18,Alpher19, Alpher20,Alpher87,Alpher88,Alpher89,Alpher90,Alpher91} has been studied from several complementary perspectives. With the introduction of QVHighlights~\cite{Alpher79}, moment retrieval and highlight detection were unified within a shared benchmark and learning framework.
Moment-DETR~\cite{Alpher79} formulated grounding as a set prediction problem, in which learnable moment queries decode candidate intervals and bipartite matching assigns predictions to ground truth moments.
This formulation has encouraged subsequent research on unified localization and saliency modeling.
Later methods\cite{Alpher08,Alpher09,Alpher32,Alpher33} improve the interaction between moment queries, video clips, and textual features, while also refining the confidence and temporal quality of decoded predictions.
Nevertheless, query-level matching and localization objectives do not necessarily evaluate whether the internal semantic structure of a query is fully supported by a candidate moment.

Subsequent studies have improved VTG from several directions.
Multimodal fusion methods~\cite{Alpher25} enhance information exchange between visual and linguistic features, while contrastive alignment~\cite{Alpher26} strengthens the separation between relevant and irrelevant video query pairs.
Multi granularity approaches~\cite{Alpher27,Alpher24} model temporal information at different resolutions or interaction levels, aiming to capture both local details and broader event context.
Other methods improve video and text representations through large scale pretraining or language models~\cite{Alpher03,Alpher30}, thereby providing stronger semantic priors for temporal localization.
These developments have substantially improved the ability of VTG models to identify query relevant content.

More recent approaches further investigate multiscale temporal reasoning~\cite{Alpher31} and structured semantic modeling~\cite{Alpher32}.
Multiscale reasoning is useful when events vary considerably in duration, while structured modeling attempts to preserve more detailed linguistic or visual information than a single sentence embedding.
However, richer temporal representations do not by themselves guarantee that the queried action is associated with the correct entity.
When an entity remains visible over an extended interval, or when similar actions occur at different locations, a model may still respond to individually relevant cues without identifying the moment in which they participate in the same event.
This motivates a more explicit treatment of compositional interaction in temporal grounding.

\subsection {Fine-Grained Cross-Modal Interaction and Assignment}

High level video understanding depends on establishing correspondences between visual evidence and linguistic expressions.
Foundation models such as CLIP~\cite{Alpher41} provide strong joint visual and textual representations, while earlier grounding methods often combine modalities through direct concatenation or shallow fusion, as exemplified by MINI-Net~\cite{Alpher77}.
These representations provide useful semantic similarity but tend to compress the query into a holistic embedding.
Such compression can preserve overall relevance while obscuring the different roles played by actions, entities, attributes, and relations in the described event.

Later approaches introduce query conditioned interaction to obtain more selective video representations.
QD-DETR~\cite{Alpher26} and CG-DETR~\cite{Alpher24}, for example, use linguistic information to guide the encoding or decoding of temporal content.
By conditioning visual features on the query, these methods reduce the influence of obviously unrelated clips and improve moment discrimination.
However, query conditioning mainly determines how strongly a clip relates to the complete query.
It does not necessarily reveal which part of the relevance originates from an action, which part originates from an entity, or whether these two sources of evidence are compositionally consistent.

Saliency guided approaches~\cite{Alpher27,Alpher31} provide another important direction.
They estimate clip relevance and use saliency information to emphasize temporally informative content.
This is particularly useful for suppressing background clips and improving the joint treatment of moment retrieval and highlight detection.
Nevertheless, a high saliency response may be caused by a persistent subject, a visually prominent object, or a frequently occurring action.
Without an explicit interaction constraint, saliency alone cannot determine whether the queried action is performed by the queried entity within the selected moment.
This distinction is important because visual prominence and event completeness are not equivalent.

Fine-grained and structured grounding methods have increasingly
explored richer linguistic organization beyond holistic sentence
representations. Token-level and local cross-modal interaction
preserve word-specific evidence, while phrase-level approaches
associate different textual constituents with corresponding visual
content. DualGround~\cite{Alpher32}, for example, performs
dual-grained alignment at the phrase and sentence levels to retain
complementary semantic granularity. Structured compositional
grounding methods~\cite{Alpher76} further model relations among
multiple semantic elements through graph-based correspondence,
providing a more expressive representation of complex event
structure.

Our focus is complementary but more specific. Rather than modeling
unrestricted phrase-level or graph-level relations, IAE-VTG targets
the action--entity binding ambiguity that directly affects temporal
localization.

\noindent\textbf{Discussion.}
Existing VTG studies have progressively improved temporal proposals, direct boundary prediction, set based decoding, multimodal fusion, saliency estimation, and fine grained alignment.
However, detecting action related and entity related evidence remains different from verifying that they participate in the same event.
Moreover, representation improvements alone cannot prevent semantically incomplete predictions from receiving positive supervision during bipartite matching.
Our IAE-VTG addresses these two aspects jointly.
FDIM disentangles action related and entity related query information, grounds them in complementary motion and appearance streams, and explicitly models their interaction to construct composition sensitive representations.
ISA further introduces binding and saliency consistency into bipartite matching, aligning the assignment criterion with the same interaction requirement used in representation learning.

\begin{figure*}[t]
    \centering
    \includegraphics[width=\textwidth]{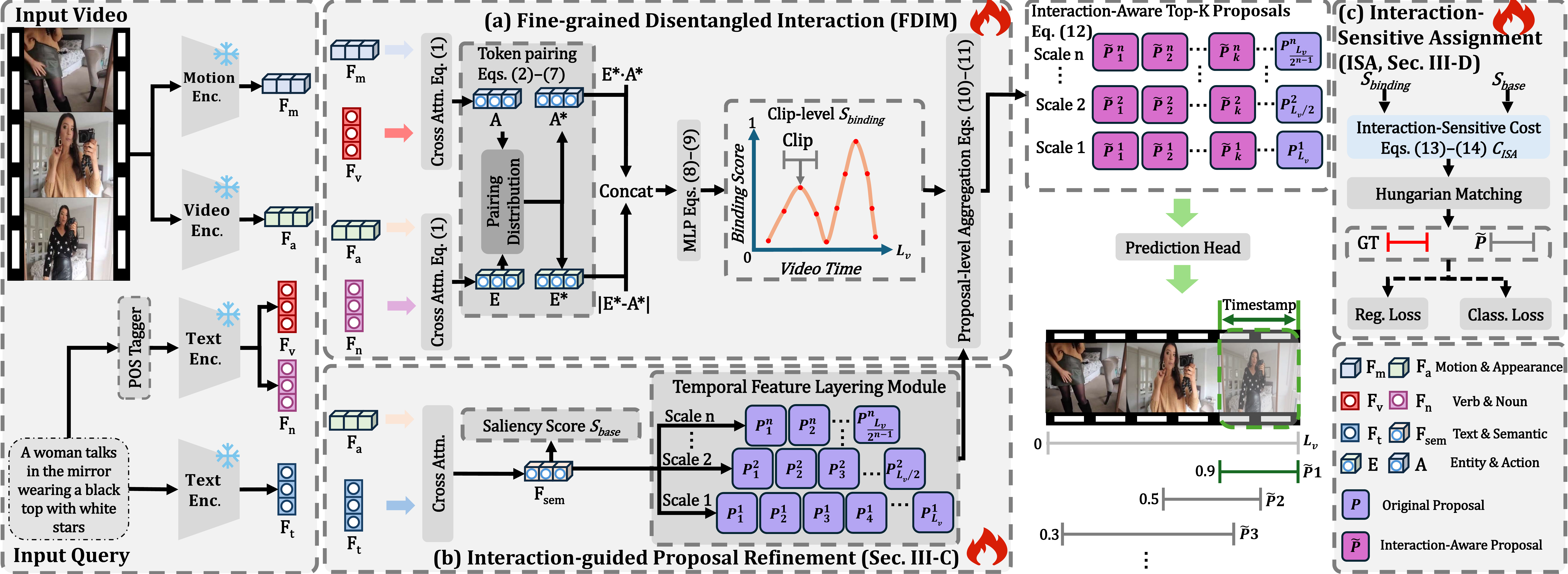}
   \caption{
    \textbf{Overview of IAE-VTG.} 
    (a) \textbf{FDIM}: Dual encoders extract motion and appearance features to yield action-aware ($\mathrm{A}$) and entity-aware ($\mathrm{E}$) representations via cross-modal attention. After token pairing, four features, $\mathrm{A}^*$, $\mathrm{E}^*$, $\mathrm{E}^* \odot \mathrm{A}^*$, and $|\mathrm{E}^* - \mathrm{A}^*|$, are concatenated to compute clip-level binding scores $S_{\text{binding}}$ via an MLP. 
    (b) \textbf{Proposal Generation}: Multi-scale proposals $P$ and saliency scores $S_{\text{base}}$ are generated from semantic features $F_{\text{sem}}$. $P$ are then modulated by aggregated $S_{\text{binding}}$ to produce interaction-refined proposals $\tilde{P}$. 
    (c) \textbf{ISA}: During training, an interaction-aware matching cost $C_{ISA}$ incorporating $S_{\text{base}}$ and $S_{\text{binding}}$ guides the label assignment for precise Video Temporal Grounding (VTG). The three stages correspond to Eqs.~\eqref{eq:role_grounding}--\eqref{eq:binding_score}, Eqs.~\eqref{eq:base_saliency}--\eqref{eq:proposal_refinement}, and Eqs.~\eqref{eq:proposal_score_pooling}--\eqref{eq:training_objective}, respectively.
    }
    \label{fig:IAE-VTG}
\end{figure*}

\section{Method}
\label{sec:method}

\subsection{System Overview}
\label{sec:overview}

Given an untrimmed video
$\mathcal{V}=\{v_t\}_{t=1}^{T}$
and a natural language query
$\mathcal{Q}=\{w_l\}_{l=1}^{L}$,
Video Temporal Grounding aims to localize the temporal interval
described by the query.
Following the underlying VTG backbone, we extract appearance and
motion features
$\mathbf{F}_a,\mathbf{F}_m\in\mathbb{R}^{T\times d}$.
The backbone predicts a set of temporal proposals
$\mathcal{P}=\{p_i\}_{i=1}^{N}$,
where each proposal
$p_i=(t_s^i,t_e^i)$
has an initial confidence $c_i$.

As illustrated in Fig.~\ref{fig:IAE-VTG}, IAE-VTG consists of
three stages built upon the underlying VTG backbone.
First, the Fine-grained Disentangled Interaction Module (FDIM),
shown in Fig.~\ref{fig:IAE-VTG}(a), grounds action and
entity related query tokens in complementary motion and appearance
streams and constructs clip-level interaction evidence
(Eqs.~\eqref{eq:role_grounding}--\eqref{eq:binding_score}).
Second, Fig.~\ref{fig:IAE-VTG}(b) shows how the resulting binding
scores are aggregated at the proposal level and used to refine
interaction-supported candidates
(Eqs.~\eqref{eq:base_saliency}--\eqref{eq:proposal_refinement}).
Finally, the Interaction-Sensitive Assignment (ISA) in
Fig.~\ref{fig:IAE-VTG}(c) incorporates the same interaction
evidence into bipartite matching during training
(Eqs.~\eqref{eq:proposal_score_pooling}--\eqref{eq:training_objective}).

Three scores are used throughout the framework.
$S_{\mathrm{base}}(t)$ denotes the clip-level relevance predicted
by the grounding backbone;
$S_{\mathrm{binding}}(t)$ measures whether the queried action and
entity are jointly supported at clip $t$;
and $c_p$ denotes the confidence of proposal $p$.
The distinction between the first two scores is important:
$S_{\mathrm{base}}$ captures general query relevance, whereas
$S_{\mathrm{binding}}$ evaluates the internal composition of the
queried event.

\subsection{Fine-grained Disentangled Interaction Module}
\label{sec:fdim}

As shown in Fig.~\ref{fig:IAE-VTG}(a), FDIM constructs
interaction evidence in three steps: role-specific grounding,
action--entity token pairing, and interaction binding.

\subsubsection{Role-specific Grounding}

We use a lightweight linguistic parser to identify noun and verb
tokens in the query.
Let
$\mathbf{X}=\{\mathbf{x}_l\}_{l=1}^{L}\in\mathbb{R}^{L\times d}$
denote the encoded query-token sequence.
Binary role masks $\mathbf{M}_n$ and $\mathbf{M}_v$ indicate the
noun and verb tokens, respectively, yielding the corresponding
role-specific token sets
$\mathbf{F}_n=\{\mathbf{x}_n\}_{n=1}^{N_n}$
and
$\mathbf{F}_v=\{\mathbf{x}_v\}_{v=1}^{N_v}$.
The complete query representation is still used by the grounding
backbone; the role-specific masks are introduced only for interaction
modeling.

Appearance and motion provide complementary evidence.
Appearance tends to preserve subjects, objects, and scene attributes,
whereas motion is more responsive to temporal changes.
We therefore obtain action-aware and entity-aware clip
representations, together with their token-level response matrices,
through
\begin{equation}
\begin{aligned}
(\mathbf{A},\mathbf{R}^{m})
&=
\operatorname{MHA}
(\mathbf{F}_{m},\mathbf{X},\mathbf{X};\mathbf{M}_{v}),\\
(\mathbf{E},\mathbf{R}^{a})
&=
\operatorname{MHA}
(\mathbf{F}_{a},\mathbf{X},\mathbf{X};\mathbf{M}_{n}),
\end{aligned}
\label{eq:role_grounding}
\end{equation}
where
$\operatorname{MHA}(\mathbf{Q},\mathbf{K},\mathbf{V};\mathbf{M})$
denotes multi-head cross-attention with the role mask $\mathbf{M}$
applied to the key/value tokens.
The resulting representations satisfy
\begin{equation*}
\mathbf{A},\mathbf{E}\in\mathbb{R}^{T\times d},
\end{equation*}
where $\mathbf{A}_t$ denotes the clip-level action-aware
representation obtained by aggregating the valid verb tokens, and
$\mathbf{E}_t$ denotes the corresponding entity-aware representation
obtained from the valid noun tokens.

The associated response matrices are
\begin{equation*}
\mathbf{R}^{m}\in\mathbb{R}^{T\times N_v},
\qquad
\mathbf{R}^{a}\in\mathbb{R}^{T\times N_n}.
\end{equation*}
Here, $R^{m}_{t,v}$ measures the response of clip $t$ to verb token
$v$, while $R^{a}_{t,n}$ measures its response to noun token $n$.
The response matrices are obtained from the softmax-normalized
cross-attention weights and averaged over attention heads.
Thus, the cross-attention stage aggregates role-specific token
features into clip-level representations while retaining token-wise
attention responses for the subsequent noun--verb pairing stage.
When a word is divided into multiple subword tokens, their features
and responses are averaged before pairing. The clip-level representations $\mathbf{A}$ and $\mathbf{E}$ serve as role-conditioned summaries of the motion and appearance streams,
whereas the subsequent token-pairing stage operates on the associated
response matrices $\mathbf{R}^{m}$ and $\mathbf{R}^{a}$ together
with the original token features. The resulting pair probabilities
are then used to construct the pair-aware representations
$\mathbf{A}^{*}$ and $\mathbf{E}^{*}$.

\subsubsection{Action--Entity Token Pairing}
\label{sec:token_pairing}

A query may contain several nouns and verbs, but only some of their
combinations describe the target event.
For efficiency, we retain the $K_n$ nouns and $K_v$ verbs with the
largest accumulated role-specific attention responses and construct
the candidate pair set
\begin{equation}
\begin{aligned}
\Omega_n
&=
\operatorname{TopK}_{n}
\left(
\sum_{t=1}^{T} R^{a}_{t,n},
K_n
\right),\\
\Omega_v
&=
\operatorname{TopK}_{v}
\left(
\sum_{t=1}^{T} R^{m}_{t,v},
K_v
\right),\\
\Omega
&=
\Omega_n \times \Omega_v .
\end{aligned}
\label{eq:pair_set}
\end{equation}
The parser only determines the token roles; the correspondence
between nouns and verbs is learned by the model.

For each clip $t$, the probability of pairing noun $n$ with verb
$v$ is defined as
\begin{equation}
P_t(n,v)
=
\frac{\exp\!\left(\ell_t(n,v)\right)}
{\displaystyle
\sum_{(n',v')\in\Omega}
\exp\!\left(\ell_t(n',v')\right)},
\qquad
(n,v)\in\Omega ,
\label{eq:pair_distribution}
\end{equation}
where the pairing logit is
\begin{equation}
\begin{aligned}
\ell_t(n,v)
={}&
\mathcal{I}_{t,n,v}
+
\lambda_g
g\!\left(
\mathbf{x}_n,
\mathbf{x}_v
\right)
\\
&+
\mu_h
h\!\left(
\mathbf{F}_{a,t},
\mathbf{F}_{m,t},
\mathbf{x}_n,
\mathbf{x}_v
\right).
\end{aligned}
\label{eq:pair_logit}
\end{equation}

For sparse interaction modeling, only the highest-scoring
noun--verb pairs according to the pairing logits are retained for
subsequent interaction aggregation.
Unless otherwise specified, all subsequent operations over
$\Omega$ are performed on this retained sparse pair set, while
we keep the notation $\Omega$ for simplicity.
In all experiments, we retain the top $K_{\mathrm{pair}}=8$
noun-verb pairs, and this setting is fixed across datasets.

The three terms in Eq.~\eqref{eq:pair_logit} provide complementary
pairing evidence.
The cross-modal prior is defined as
\begin{equation}
\mathcal{I}_{t,n,v}
=
\log\!\left(R^{a}_{t,n}+\epsilon\right)
+
\log\!\left(R^{m}_{t,v}+\epsilon\right),
\label{eq:cross_modal_prior}
\end{equation}
which favors noun--verb pairs whose appearance-related noun response
and motion-related verb response are simultaneously strong at clip
$t$.
The function $g(\cdot)$ evaluates language-level compatibility
between noun token $\mathbf{x}_n$ and verb token $\mathbf{x}_v$,
whereas $h(\cdot)$ evaluates their compatibility under the local
appearance and motion context
$(\mathbf{F}_{a,t},\mathbf{F}_{m,t})$.
Both functions are implemented as two-layer MLPs with scalar outputs.
Their inputs combine the corresponding original features with
element-wise products and absolute differences, as illustrated in
Fig.~\ref{fig:IAE-VTG}.

We obtain noun and verb marginal probabilities from the learned
pair distribution:
\begin{equation}
P_t^{n}(n)
=
\sum_{v:(n,v)\in\Omega}P_t(n,v),
\qquad
P_t^{v}(v)
=
\sum_{n:(n,v)\in\Omega}P_t(n,v).
\label{eq:pair_marginals}
\end{equation}

The pair-aware action and entity representations are then obtained
by weighting the corresponding role-specific token features:
\begin{equation}
\mathbf{A}^{*}_t
=
\sum_{v\in\Omega_v}
P_t^{v}(v)\mathbf{x}_v,
\qquad
\mathbf{E}^{*}_t
=
\sum_{n\in\Omega_n}
P_t^{n}(n)\mathbf{x}_n.
\label{eq:paired_features}
\end{equation}
These representations summarize the verb and noun tokens whose
pairing is most strongly supported by the visual content at clip $t$.

\subsubsection{Interaction Binding}

The paired features are combined into
\begin{equation}
\begin{aligned}
\mathbf{z}_t
=
\operatorname{Concat}\big(
&\mathbf{A}^{*}_t,
\mathbf{E}^{*}_t,
\mathbf{A}^{*}_t\odot\mathbf{E}^{*}_t,\\
&|\mathbf{A}^{*}_t-\mathbf{E}^{*}_t|
\big),
\end{aligned}
\label{eq:binding_feature}
\end{equation}
\begin{equation}
S_{\mathrm{binding}}(t)
=
\sigma\!\left(
\phi_{\mathrm{bind}}(\mathbf{z}_t)
\right).
\label{eq:binding_score}
\end{equation}
The product term captures coactivation between action and entity
evidence, while the absolute difference measures their
disagreement.
The MLP $\phi_{\mathrm{bind}}$ maps this representation to the
clip-level binding score.
A high score indicates that the queried action and entity are
jointly supported at the corresponding temporal location.

\subsection{Interaction-guided Proposal Refinement}
\label{sec:proposal_refinement}

As illustrated in Fig.~\ref{fig:IAE-VTG}(b), we retain the
multi-scale proposal generation mechanism of the underlying VTG
backbone and inject the FDIM binding evidence only during
proposal-level refinement.

Given holistic semantic features $\mathbf{F}_{\mathrm{sem}}$, the
backbone constructs a temporal feature pyramid, predicts candidate
moments, and produces the baseline saliency
\begin{equation}
S_{\mathrm{base}}
=
\sigma\!\left(
\phi_{\mathrm{sem}}(\mathbf{F}_{\mathrm{sem}})
\right),
\label{eq:base_saliency}
\end{equation}
where $\phi_{\mathrm{sem}}$ is the original saliency head.

For proposal $p$ with temporal span $\mathcal{I}_p$, its interaction
support is obtained by averaging the binding scores within the
proposal:
\begin{equation}
s_p
=
\frac{1}{|\mathcal{I}_p|}
\sum_{t\in\mathcal{I}_p}S_{\mathrm{binding}}(t).
\label{eq:proposal_binding}
\end{equation}

We apply interaction modulation to the last $L_c$ levels of the
temporal pyramid, where proposals cover broader temporal contexts.
For each active level $\ell$, let $\mathcal{P}_{\ell}$ denote the
corresponding proposal set. We select the $K_p^{(\ell)}$ proposals
with the largest interaction support $s_p$, where $K_p^{(\ell)}$ is
defined as a fixed proportion of the proposals at that level.
Their confidence is refined by
\begin{equation}
\widetilde{c}_p
=
\begin{cases}
c_p(1+\alpha s_p),
&
p\in
\operatorname{TopK}_{q\in\mathcal{P}_{\ell}}
\left(s_q,K_p^{(\ell)}\right),\\
c_p,
&
\text{otherwise},
\end{cases}
\label{eq:proposal_refinement}
\end{equation}
where $\alpha$ controls the contribution of interaction evidence.
The residual form preserves the original proposal confidence while
increasing the scores of candidates with stronger action--entity
interaction support.
The refined confidence is used during both training and inference.
In all experiments, interaction refinement is applied only to the
coarsest temporal-pyramid level, i.e., $L_c=1$. For an active
level $\ell$ containing $N_\ell$ proposals, we set
$K_p^{(\ell)}=\max(1,\lfloor0.2N_\ell\rfloor)$, corresponding to
the top $20\%$ proposals ranked by interaction support. This
selection rule is fixed across datasets.

\subsection{Interaction-Sensitive Assignment and Optimization}
\label{sec:isa}
Figure~\ref{fig:IAE-VTG}(c) illustrates the training-time ISA
stage.
Proposal refinement affects prediction confidence, but it does not
determine which proposals receive positive supervision.
In DETR-based grounding models, this decision is made through
bipartite matching.
A temporally plausible proposal can therefore be selected even when
its semantic evidence is incomplete.
ISA addresses this issue by incorporating interaction consistency into
the matching cost.

For each proposal $p$, we first aggregate the backbone saliency and
interaction-binding scores over its temporal span $I_p$:
\begin{equation}
\begin{aligned}
\bar{S}_{\mathrm{base}}(p)
&=
\frac{1}{|I_p|}
\sum_{t\in I_p}
S_{\mathrm{base}}(t),\\
\bar{S}_{\mathrm{binding}}(p)
&=
\frac{1}{|I_p|}
\sum_{t\in I_p}
S_{\mathrm{binding}}(t).
\end{aligned}
\label{eq:proposal_score_pooling}
\end{equation}

The interaction-sensitive matching cost is then defined as
\begin{equation}
\begin{aligned}
\mathcal{C}_{\mathrm{ISA}}(p,g)
={}&
\mathcal{C}_{\mathrm{base}}(p,g) \\
&+
\beta\,\mathrm{IoU}(p,g)
\left|
\bar{S}_{\mathrm{binding}}(p)
-
\bar{S}_{\mathrm{base}}(p)
\right|.
\end{aligned}
\label{eq:isa_cost}
\end{equation}
where $\mathcal{C}_{\mathrm{base}}$ denotes the original temporal
matching cost and $\beta$ controls the contribution of interaction
consistency.
The additional term measures the discrepancy between holistic query
relevance and action--entity binding evidence.
Its IoU weighting focuses the semantic consistency constraint on
temporally plausible proposals, while limiting the influence of
candidates that are already poorly aligned with the ground-truth
moment.
Consequently, proposals with strong temporal overlap but inconsistent
interaction evidence receive a larger assignment cost.

The standard Hungarian algorithm is applied using
$\mathcal{C}_{\mathrm{ISA}}$.
The matching operation remains discrete and is not differentiated;
ISA affects training by changing which proposals are selected as
positive targets.
It introduces no additional operation at inference time.

The backbone retains its original localization, classification, and
saliency losses:
\begin{equation}
\mathcal{L}
=
\lambda_{\mathrm{loc}}\mathcal{L}_{\mathrm{loc}}
+
\lambda_{\mathrm{cls}}
\mathcal{L}_{\mathrm{cls}}(\widetilde{\mathbf{c}})
+
\lambda_{\mathrm{sal}}\mathcal{L}_{\mathrm{sal}}.
\label{eq:training_objective}
\end{equation}

Because $\widetilde{c}_p$ depends differentiably on
$S_{\mathrm{binding}}$, the classification loss provides a direct
optimization path for FDIM.
ISA further influences learning through interaction-sensitive target
assignment without propagating gradients through Hungarian matching.
Accordingly, interaction evidence affects optimization through two
complementary paths: differentiable proposal refinement and
interaction-sensitive supervision assignment.

Queries without an identified noun or verb use the highest-attended
content token as a fallback. The complete architectural and training
settings are specified in Sec.~\ref{sec:implementation}.

\section{Experiments}

\subsection{Implementation Details}
\label{sec:implementation}

\noindent\textbf{Dataset.} 
Following FlashVTG~\cite{Alpher31}, we adopt the same data preprocessing pipeline and train/val/test splits. 
Experiments are conducted on three VTG benchmarks: QVHighlights~\cite{Alpher79}, Charades-STA~\cite{Alpher14}, and TACoS~\cite{Alpher42}. 
QVHighlights~\cite{Alpher79} serves as the primary benchmark with full comparisons, while Charades-STA and TACoS are used to evaluate moment retrieval performance in daily-activity and cooking scenarios.

\noindent\textbf{Metrics.}
We follow the evaluation protocol of FlashVTG~\cite{Alpher31}. 
For moment retrieval, we report R1@X ($X\in\{0.3,0.5,0.7\}$), mAP (averaged over IoU thresholds from 0.5 to 0.95 with step size 0.05 following COCO-style evaluation~\cite{Alpher44}), mIoU, and mAP@0.5/0.75 for consistency with prior work. 
For highlight detection, we report mAP.

\noindent\textbf{Hyperparameters and Architectural Settings.}
IAE-VTG is implemented in PyTorch and trained end-to-end using
the AdamW optimizer on a single NVIDIA A6000 GPU.
Unless otherwise specified, we follow the training configuration
of the underlying FlashVTG backbone.

The proposal-modulation coefficient and ISA interaction weight are
set to $\alpha=0.6$ and $\beta=0.8$, respectively, based on the
QVHighlights validation set. For FDIM, the language-compatibility
weight, visual-context compatibility weight, and pairing temperature
are fixed to $\lambda_g=0.7$, $\mu_h=0.4$, and $\tau=1.0$.
We retain at most $K_n=12$ noun tokens and $K_v=12$ verb tokens,
followed by the top $K_{\mathrm{pair}}=8$ noun--verb pairs.

Interaction-guided proposal refinement is applied only to the
coarsest temporal-pyramid level ($L_c=1$), where the top $20\%$
proposals ranked by interaction support are modulated.
The temporal pyramid contains five levels with strides
$(1,2,4,8,16)$.

All interaction representations use the same $d=256$ dimensional
latent space as the grounding backbone. The role-specific
cross-attention modules use four attention heads, and the FDIM MLPs
use a hidden dimension of 512. The log-prior stabilizer is set to
$\epsilon=10^{-8}$. All of these architectural and selection
settings are kept fixed across QVHighlights, Charades-STA, and TACoS.

The noun--verb decomposition is used only by the interaction branch;
the complete sentence representation remains unchanged in the
original proposal-generation and saliency pathways.
Sensitivity analyses of $\alpha$ and $\beta$ are shown in
Figs.~\ref{fig:line} and~\ref{fig:radar}.
Extended comparisons with large-scale and detector-enhanced VTG
models are presented in Sec.~\ref{sec:extended_comparisons}.

\noindent\textbf{Efficiency.}
IAE-VTG increases the parameter count from 11.81M to 13.26M
and runs at 139 FPS, compared with 168 FPS for the baseline.
ISA is training-only and introduces no additional inference-time
operation.

\begin{table*}[t]
\centering
\caption{
Comparison with state-of-the-art methods on QVHighlights~\cite{Alpher79}
under the test and validation splits. Best results are shown in \textbf{bold}, and
second-best results are \underline{underlined}.
}
\label{tab:sota_comparison}

\small
\setlength{\tabcolsep}{4.5pt}
\renewcommand{\arraystretch}{0.95}

\begin{tabular}{l|ccccc|ccccc}
\toprule
\multirow{3}{*}{\textbf{Method}}
& \multicolumn{5}{c|}{\textbf{Test Set}}
& \multicolumn{5}{c}{\textbf{Validation Set}} \\
\cmidrule(lr){2-6}
\cmidrule(lr){7-11}

& \multicolumn{2}{c}{\textbf{R1}}
& \multicolumn{3}{c|}{\textbf{mAP}}
& \multicolumn{2}{c}{\textbf{R1}}
& \multicolumn{3}{c}{\textbf{mAP}} \\
\cmidrule(lr){2-3}
\cmidrule(lr){4-6}
\cmidrule(lr){7-8}
\cmidrule(lr){9-11}

& @0.5 & @0.7
& @0.5 & @0.75 & Avg.
& @0.5 & @0.7
& @0.5 & @0.75 & Avg. \\
\midrule

TaskWeave~\cite{Alpher33}
{\footnotesize\color{gray} CVPR'24}
& -- & -- & -- & -- & --
& 64.26 & 50.06 & 65.39 & 46.47 & 45.38 \\

CG-DETR~\cite{Alpher24}
{\footnotesize\color{gray} arXiv'23}
& 65.43 & 48.38 & 64.51 & 42.77 & 42.86
& 67.35 & 52.06 & 65.57 & 45.73 & 44.93 \\

UVCOM~\cite{Alpher27}
{\footnotesize\color{gray} CVPR'24}
& 63.55 & 47.47 & 63.37 & 42.67 & 43.18
& 65.10 & 51.81 & -- & -- & 45.79 \\

LLMEPET~\cite{Alpher03}
{\footnotesize\color{gray} ACM MM'24}
& 66.73 & 49.94 & 65.76 & 43.91 & 44.05
& 66.58 & 51.10 & -- & -- & 46.24 \\

R$^2$-Tuning~\cite{Alpher30}
{\footnotesize\color{gray} ECCV'24}
& 68.03 & 49.35 & 69.04 & 47.56 & 46.17
& 68.71 & 52.06 & -- & -- & 47.59 \\

FlashVTG~\cite{Alpher31}
{\footnotesize\color{gray} WACV'25}
& 70.69 & 53.96 & 72.33 & 53.85 & 52.00
& 73.10 & 57.29 & 72.75 & 54.33 & 52.84 \\

DualGround~\cite{Alpher32}
{\footnotesize\color{gray} NeurIPS'25}
& \underline{71.87}
& \textbf{56.94}
& \underline{72.41}
& \underline{54.38}
& \underline{52.73}
& \underline{73.48}
& \textbf{58.97}
& \underline{72.99}
& \underline{56.35}
& \underline{53.26} \\

KDA~\cite{Alpher35}
{\footnotesize\color{gray} ICCV'25}
& 66.70 & 50.88 & 67.57 & 46.31 & 45.67
& 69.11 & 53.46 & 68.17 & 48.04 & 47.41 \\

\midrule

\textbf{IAE-VTG(Ours)}
& \textbf{71.92}
& \underline{55.90}
& \textbf{73.51}
& \textbf{55.31}
& \textbf{52.91}
& \textbf{73.81}
& \underline{58.52}
& \textbf{73.49}
& \textbf{57.16}
& \textbf{54.47} \\

\bottomrule
\end{tabular}
\end{table*}

\begin{table}[t]
\centering
\caption{
Comparison on the Charades-STA~\cite{Alpher14} test set.
``SF+C'' denotes SlowFast R-50~\cite{Alpher40} combined with
CLIP-B/32~\cite{Alpher41}, while ``IV2'' denotes
InternVideo2-6B~\cite{Alpher01}.
Best results are shown in \textbf{bold}, and second-best results are
\underline{underlined}.
}
\label{tab:charades_sta_results}

\small
\setlength{\tabcolsep}{5.0pt}
\renewcommand{\arraystretch}{1.10}

\begin{tabular}{lccc}
\toprule
\textbf{Method}
& \textbf{Backbone}
& \textbf{R1@0.5}
& \textbf{R1@0.7} \\
\midrule

2D-TAN~\cite{Alpher36}
& SF+C & 46.02 & 27.50 \\

VSLNet~\cite{Alpher37}
& SF+C & 42.69 & 24.14 \\

Moment-DETR~\cite{Alpher79}
& SF+C & 52.07 & 30.59 \\

QD-DETR~\cite{Alpher26}
& SF+C & 57.31 & 32.55 \\

UniVTG~\cite{Alpher05}
& SF+C & 58.01 & 35.65 \\

TR-DETR~\cite{Alpher39}
& SF+C & 57.61 & 33.52 \\

LLMEPET~\cite{Alpher03}
& SF+C & -- & 36.49 \\

CG-DETR~\cite{Alpher24}
& SF+C & \underline{58.44} & 36.34 \\

FlashVTG~\cite{Alpher31}
& SF+C & 57.58 & \underline{37.31} \\

\textbf{IAE-VTG (Ours)}
& SF+C & \textbf{58.63} & \textbf{37.58} \\

\midrule

FlashVTG~\cite{Alpher31}
& IV2 & \underline{70.32} & \underline{49.87} \\

\textbf{IAE-VTG (Ours)}
& IV2 & \textbf{71.42} & \textbf{50.03} \\

\bottomrule
\end{tabular}
\end{table}

\begin{table}[t]
\centering
\caption{
Comparison on TACoS~\cite{Alpher42}.
All methods use SlowFast~\cite{Alpher40} and
CLIP~\cite{Alpher41} as visual and textual backbones.
Best results are shown in \textbf{bold}, and second-best results are
\underline{underlined}.
}
\label{tab:tacos}

\small
\setlength{\tabcolsep}{5.0pt}
\renewcommand{\arraystretch}{1.12}

\begin{tabular}{lcccc}
\toprule
\textbf{Method}
& \textbf{R1@0.3}
& \textbf{R1@0.5}
& \textbf{R1@0.7}
& \textbf{mIoU} \\
\midrule

2D-TAN~\cite{Alpher36}
& 40.01 & 27.99 & 12.92 & 27.22 \\

VSLNet~\cite{Alpher37}
& 35.54 & 23.54 & 13.15 & 24.99 \\

Moment-DETR~\cite{Alpher79}
& 37.97 & 24.67 & 11.97 & 25.49 \\

UniVTG~\cite{Alpher05}
& 51.44 & 34.97 & 17.35 & 33.60 \\

CG-DETR~\cite{Alpher24}
& 52.23 & 39.61 & 22.23 & 36.48 \\

R$^2$-Tuning~\cite{Alpher30}
& 49.71 & 38.72 & \underline{25.12} & 35.92 \\

LLMEPET~\cite{Alpher03}
& 52.73 & -- & 22.78 & 36.55 \\

FlashVTG~\cite{Alpher31}
& \underline{53.71}
& \underline{41.76}
& 24.74
& \underline{37.61} \\

\midrule

\textbf{IAE-VTG (Ours)}
& \textbf{54.21}
& \textbf{42.66}
& \textbf{26.57}
& \textbf{39.25} \\

\bottomrule
\end{tabular}
\end{table}

We compare IAE-VTG with recent state-of-the-art VTG systems across MR/HD benchmarks, including proposal-free DETR-style baselines and their variants {Moment-DETR}~\cite{Alpher79}, {QD-DETR}~\cite{Alpher26}, {TR-DETR}~\cite{Alpher39}, {UniVTG}~\cite{Alpher05}, {CG-DETR}~\cite{Alpher24}, as well as stronger recent models such as {FlashVTG}~\cite{Alpher31}, {DualGround}~\cite{Alpher32}, {KDA}~\cite{Alpher35}, {LLMEPET}~\cite{Alpher03}, {$R^2$-Tuning}~\cite{Alpher30}, UVCOM~\cite{Alpher27}, and {Task-Weave}~\cite{Alpher33}. 
For MR datasets, we additionally include classical baselines {2D-TAN}~\cite{Alpher36} and {VSLNet}~\cite{Alpher37}. 

\begin{table*}[t]
\centering
\caption{
Ablation study of FDIM and ISA on QVHighlights~\cite{Alpher79}.
$\checkmark$ indicates that the corresponding component is enabled.
Results are reported on both the test and validation splits.
Best results are shown in \textbf{bold}.
}
\label{tab:ablation_main}

\small
\setlength{\tabcolsep}{5pt}
\renewcommand{\arraystretch}{0.80}

\begin{tabular}{cccccccccccc}
\toprule
\multirow{2}{*}{\textbf{FDIM}}
& \multirow{2}{*}{\textbf{ISA}}
& \multicolumn{5}{c}{\textbf{Test Set}}
& \multicolumn{5}{c}{\textbf{Validation Set}} \\
\cmidrule(lr){3-7}
\cmidrule(lr){8-12}

&
& \textbf{R1@0.5}
& \textbf{R1@0.7}
& \textbf{mAP@0.5}
& \textbf{mAP@0.75}
& \textbf{Avg.}
& \textbf{R1@0.5}
& \textbf{R1@0.7}
& \textbf{mAP@0.5}
& \textbf{mAP@0.75}
& \textbf{Avg.} \\
\midrule

$\times$
& $\times$
& 70.69
& 53.96
& 72.33
& 53.85
& 52.00
& 71.48
& 56.06
& 72.37
& 55.03
& 52.61 \\

$\checkmark$
& $\times$
& 71.27
& 54.93
& 73.20
& 54.56
& 52.17
& \textbf{73.42}
& 57.23
& 73.16
& 54.83
& 52.71 \\

$\checkmark$
& $\checkmark$
& \textbf{71.92}
& \textbf{55.90}
& \textbf{73.51}
& \textbf{55.31}
& \textbf{52.91}
& \textbf{73.81}
& \textbf{58.52}
& \textbf{73.49}
& \textbf{57.16}
& \textbf{54.47} \\

\bottomrule
\end{tabular}
\end{table*}

\subsection{Quantitative Comparison}
\noindent\textbf{Performance on QVHighlights.} 
As shown in Table~\ref{tab:sota_comparison}, IAE-VTG achieves state-of-the-art or competitive performance across all metrics. On the Test set, we achieve 52.91\% Average mAP, exceeding FlashVTG~\cite{Alpher31} (+0.91\%). Notably, the gains on stricter metrics such as mAP@0.75
demonstrate improved high-precision localization.
The contribution of ISA is examined separately in the ablation study.

\noindent\textbf{Performance on Charades-STA and TACoS.} 
Evaluations on Charades-STA (Table~\ref{tab:charades_sta_results}) and TACoS (Table~\ref{tab:tacos}) further demonstrate generalizability. On Charades-STA, IAE-VTG reaches 37.58\% R1@0.7 (SF+C) and 50.03\% R1@0.7 (IV2), demonstrating generalization across different visual feature settings. On the long-form TACoS dataset, our model achieves 39.25\% mIoU and 26.57\% R1@0.7, outperforming FlashVTG (+1.64\% and +1.83\%, respectively). These results further demonstrate that IAE-VTG generalizes to
long-form videos with fine-grained procedural actions.

\noindent\textbf{Summary.}
Across all benchmarks, IAE-VTG consistently improves strict
localization metrics across diverse video domains and feature settings.
These results highlight that explicitly modeling the internal
consistency between motion and appearance provides complementary
benefits beyond conventional holistic feature fusion.
Further comparisons with large-scale and detector-enhanced VTG
architectures are presented in Sec.~\ref{sec:extended_comparisons}.

\begin{table}[t]
\centering
\caption{
Feature ablation on the QVHighlights~\cite{Alpher79} validation set.
Baseline+Dual augments the baseline with SlowFast motion features.
Best results are shown in \textbf{bold}.
}
\label{tab:feature_ablation}

\small
\setlength{\tabcolsep}{5.0pt}
\renewcommand{\arraystretch}{0.80}

\begin{tabular}{lccc}
\toprule
\textbf{Method}
& \textbf{R1@0.7}
& \textbf{mAP@0.75}
& \textbf{Avg. mAP} \\
\midrule

Baseline
& 56.06
& 55.03
& 52.61 \\

Baseline+Dual
& 57.03
& 56.29
& 53.33 \\

\textbf{IAE-VTG (Ours)}
& \textbf{58.52}
& \textbf{57.16}
& \textbf{54.47} \\

\bottomrule
\end{tabular}
\end{table}

\begin{table}[t]
\centering
\caption{
Performance on the NA-VMR task.
RA-ID and RA-OOD denote rejection accuracy for in-domain and
out-of-domain negative queries, respectively.
}
\label{tab:na_vmr}

\small
\setlength{\tabcolsep}{4.2pt}
\renewcommand{\arraystretch}{1.12}

\begin{tabular}{lcccc}
\toprule
\textbf{Method}
& \textbf{R1@0.5}
& \textbf{R1@0.7}
& \textbf{RA-ID}
& \textbf{RA-OOD} \\
\midrule

Baseline
& 68.00
& 55.87
& 52.84
& 66.84 \\

\textbf{IAE-VTG (Ours)}
& \textbf{68.90}
& \textbf{57.55}
& \textbf{68.97}
& \textbf{74.06} \\

\midrule

$\Delta$
& \textbf{+0.90}
& \textbf{+1.68}
& \textbf{+16.13}
& \textbf{+7.22} \\

\bottomrule
\end{tabular}
\end{table}

\begin{figure*}[t]
    \centering
    \includegraphics[width=\textwidth]{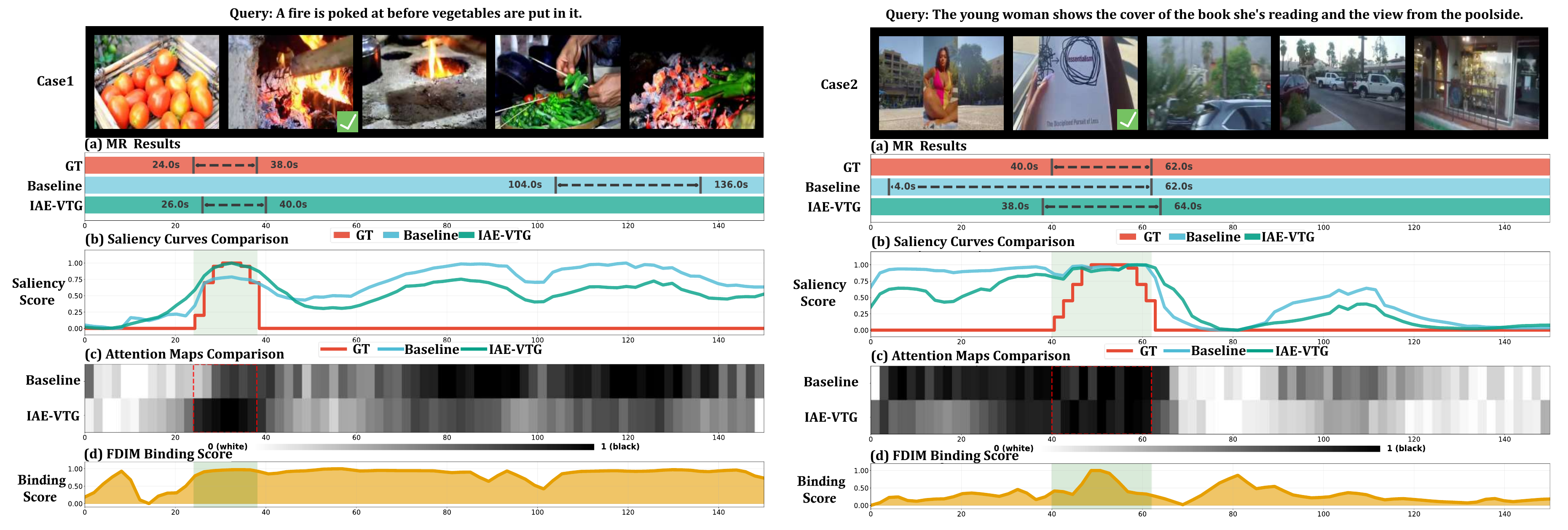}
   \caption{
Qualitative comparison on two representative temporal-ambiguity cases.
\textbf{Left:} a sequential-action case, where the baseline is distracted
by a later event while IAE-VTG localizes the queried interaction.
\textbf{Right:} an entity-persistence case, where the baseline responds
broadly to a visually persistent entity while IAE-VTG concentrates on
the target interval.
For each case, (a) compares the predicted temporal moments with the
ground truth (GT), (b) shows the temporal saliency responses,
(c) visualizes the attention maps, and (d) reports the FDIM binding
score $S_{\mathrm{binding}}$.
Together, the two examples illustrate how interaction-aware modeling
suppresses temporally plausible but compositionally inconsistent
responses.
}
    \label{fig:qualitative_cases}
\end{figure*}

\begin{figure*}[t]
    \centering
    \includegraphics[width=\textwidth]{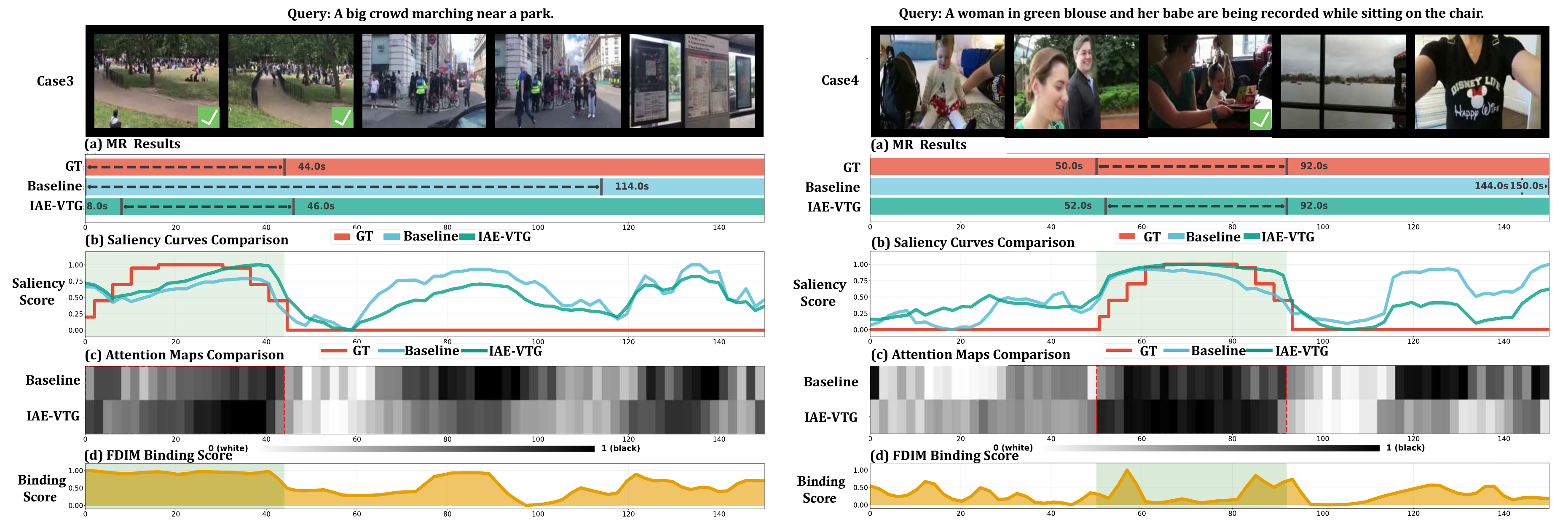}
   \caption{
Qualitative comparison on two representative temporal-ambiguity
cases.
\textbf{Left:} a temporal over-extension case, where persistent
crowd and scene evidence causes the baseline to respond far beyond
the queried ``marching'' event, whereas IAE-VTG produces a compact
prediction closely aligned with the ground truth.
\textbf{Right:} a temporally separated distractor case, where the
baseline is attracted by a late visually salient interval, while
IAE-VTG localizes the target action--entity interaction.
For each case, (a) compares the predicted temporal moments with the
ground truth (GT), (b) shows the temporal saliency responses,
(c) visualizes the attention maps, and (d) reports the FDIM binding
score $S_{\mathrm{binding}}$.
Together, the examples illustrate how interaction-aware modeling
suppresses both persistent appearance responses and temporally
separated distractors.
}
    \label{fig:qualitative_cases2}
\end{figure*}

\begin{figure*}[t]
    \centering
    \includegraphics[width=\textwidth]{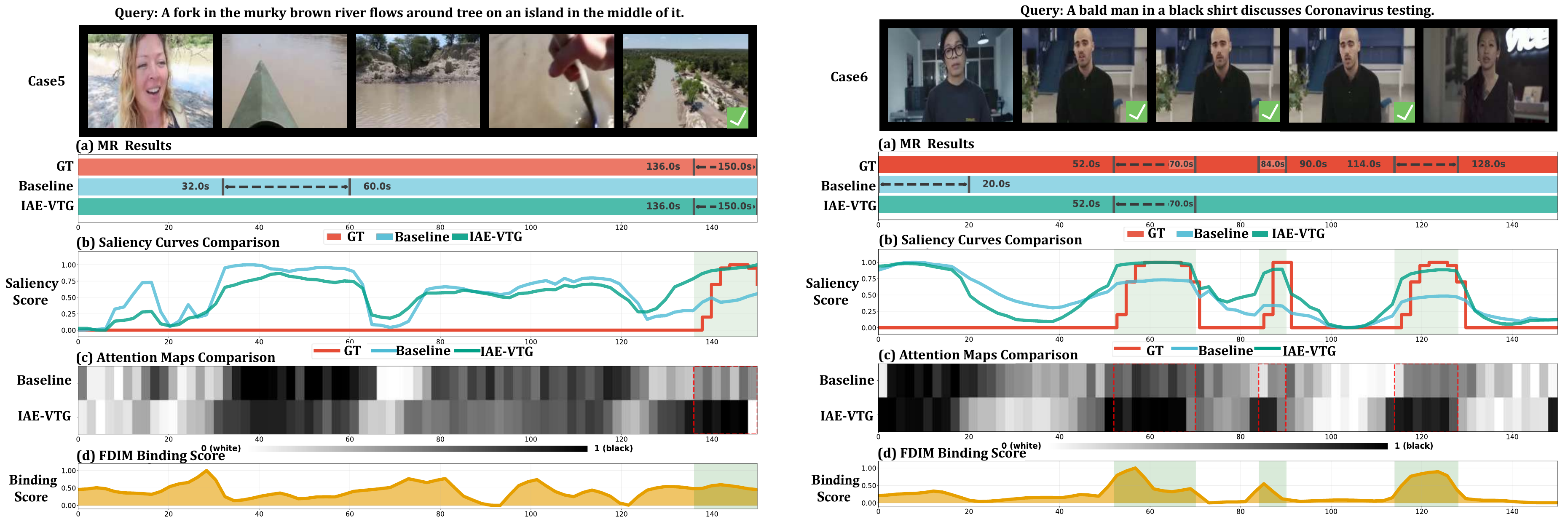}
    \caption{
    Qualitative comparison on two additional challenging
    temporal-ambiguity cases.
    \textbf{Left (Case 5):} an appearance-dominant ambiguity case,
    where the baseline localizes an earlier visually plausible
    interval, whereas IAE-VTG shifts its prediction toward the
    annotated event near the end of the video.
    \textbf{Right (Case 6):} a repeated-event ambiguity case with
    multiple temporally separated relevant intervals, where the
    baseline is dominated by an early entity-driven response while
    IAE-VTG places greater emphasis on the action-consistent regions.
    For each case, (a) compares the predicted temporal moments with
    the ground truth (GT), (b) shows the temporal saliency responses,
    (c) visualizes the attention maps, and (d) reports the FDIM
    binding score $S_{\mathrm{binding}}$.
    Together, these examples further demonstrate that explicit
    action--entity interaction evidence helps suppress appearance
    shortcuts and distinguish temporally competing events.
    }
    \label{fig:qualitative_cases3}
\end{figure*}

\subsection{Qualitative Comparison}
\label{sec:qualitative}

\textbf{Qualitative Analysis.}
Figures~\ref{fig:qualitative_cases},
\ref{fig:qualitative_cases2}, and
\ref{fig:qualitative_cases3}
present six representative examples covering complementary forms of
temporal ambiguity.
For each case, we visualize the predicted moments, temporal saliency
responses, attention maps, and FDIM binding scores.
These examples illustrate how conventional relevance-based grounding
can be affected by temporally separated actions, persistent
appearance evidence, and competing visually plausible events, whereas
IAE-VTG produces more interaction-consistent temporal responses.

\textbf{Case 1: Sequential-action ambiguity.}
In the left example of Fig.~\ref{fig:qualitative_cases}, the query
describes ``a fire is poked at before vegetables are put in it.''
The baseline is distracted by a later event and predicts an incorrect
interval around $104$--$136$\,s, although the queried interaction
occurs much earlier.
IAE-VTG instead localizes approximately $26$--$40$\,s, closely
matching the annotated moment.
The saliency and attention responses become more concentrated around
the target event, while the FDIM binding response in row (d) provides
additional evidence for the interval in which the queried action and
entity are jointly supported.
This example shows that interaction-aware evidence can distinguish
the target event from a temporally separated action with related
visual content.

\textbf{Case 2: Entity-persistence ambiguity.}
The right example of Fig.~\ref{fig:qualitative_cases} illustrates a
different failure mode.
The queried woman remains visually salient for an extended period,
causing the baseline to produce an overly broad prediction of
approximately $4$--$62$\,s.
In contrast, IAE-VTG predicts a substantially more compact interval
of approximately $38$--$64$\,s, closely aligned with the
ground-truth moment at $40$--$62$\,s.
The corresponding saliency and attention responses show that the
baseline remains sensitive to persistent appearance evidence, whereas
IAE-VTG emphasizes the temporal region in which the queried action
and entity are jointly supported.
The binding response further provides a selective interaction cue
around the target region.

\textbf{Case 3: Temporal over-extension.}
In the left example of Fig.~\ref{fig:qualitative_cases2}, the query
describes ``a big crowd marching near a park.''
The annotated event occupies approximately the first $44$\,s of the
video.
Although the baseline identifies the relevant early content, its
prediction extends to approximately $114.5$\,s, substantially beyond
the ground-truth boundary.
This behavior is consistent with crowd- and scene-level appearance
evidence remaining visually relevant after the queried marching event
has ended.
IAE-VTG instead produces a compact prediction of approximately
$0$--$46$\,s.
The more selective saliency and attention responses, together with
the interaction signal in row (d), help distinguish persistent visual
context from the temporally bounded action.

\textbf{Case 4: Temporally separated distractor.}
The right example of Fig.~\ref{fig:qualitative_cases2} provides a
more extreme distractor case.
For the query ``a woman in green blouse and her babe are being
recorded while sitting on the chair,'' the ground-truth interaction
spans approximately $50$--$92$\,s.
The baseline is attracted by a late visually salient region near the
end of the video, despite its temporal inconsistency with the queried
event.
IAE-VTG instead localizes approximately $52$--$92$\,s, closely
following the ground-truth boundaries.
Its saliency and attention responses are concentrated around the
target interval, while the binding signal provides complementary
evidence for the relevant action--entity interaction.

\textbf{Case 5: Appearance-dominant ambiguity.}
In the left example of Fig.~\ref{fig:qualitative_cases3}, the query
describes a fork in a murky river flowing around a tree on an island.
The baseline responds strongly to an earlier visually plausible river
segment, although the annotated event occurs near the end of the
video.
IAE-VTG instead shifts its prediction toward the ground-truth
interval.
This example shows that visually similar scene content can create a
strong appearance shortcut even when it occurs at the wrong temporal
location.
The interaction-aware response provides a more selective cue for
identifying when the queried visual configuration and action are
jointly supported.

\textbf{Case 6: Repeated-event ambiguity.}
The right example of Fig.~\ref{fig:qualitative_cases3} contains
multiple temporally separated relevant intervals for the query
describing a man discussing Coronavirus testing.
The baseline is dominated by an early response that does not
correspond to the annotated events, whereas IAE-VTG places greater
emphasis on the later action-relevant regions.
The corresponding saliency and attention responses become more
consistent with the annotated intervals, while the FDIM binding
signal exhibits stronger responses around the interaction-rich
regions.
This example indicates that interaction evidence is also useful when
the queried event occurs repeatedly rather than within a single
isolated interval.

\textbf{Discussion.}
Together, the six examples reveal three recurring sources of
spurious temporal grounding.
First, Cases 1, 4, and 6 show that temporally separated or competing
events can attract the baseline despite incomplete agreement with the
queried interaction.
Second, Cases 2 and 3 demonstrate that persistent entity or scene
evidence can cause predictions to extend beyond the actual action
interval.
Third, Case 5 illustrates an appearance-dominant shortcut in which
visually plausible content is localized at an incorrect temporal
position.
Across these cases, IAE-VTG mitigates the ambiguity by complementing
holistic video--text relevance with explicit action--entity
interaction evidence.
FDIM provides composition-sensitive temporal cues, while ISA
encourages proposals consistent with the same interaction evidence to
receive more reliable supervision during training.
These qualitative observations are consistent with the quantitative,
perturbation, and ablation results reported below.
    

\begin{table}[t]
\centering
\caption{
Semantic interaction perturbation analysis on
QVHighlights~\cite{Alpher79}.
}
\label{tab:semantic_swap}

\small
\renewcommand{\arraystretch}{0.80}

\begin{tabular*}{\columnwidth}
{@{\extracolsep{\fill}}lcccc}
\toprule
\textbf{Type}
& \textbf{N}
& \textbf{IoU Drop}
& \textbf{Conf. Drop}
& \textbf{Rej. (\%)} \\
\midrule

Verb
& 521
& 0.416
& 0.272
& 78.9 \\

Object
& 869
& \textbf{0.447}
& \textbf{0.275}
& \textbf{81.2} \\

Relation
& 488
& 0.410
& 0.257
& 77.3 \\

\bottomrule
\end{tabular*}
\end{table}









\begin{figure}[t]
    \centering
    \includegraphics[width=\linewidth]{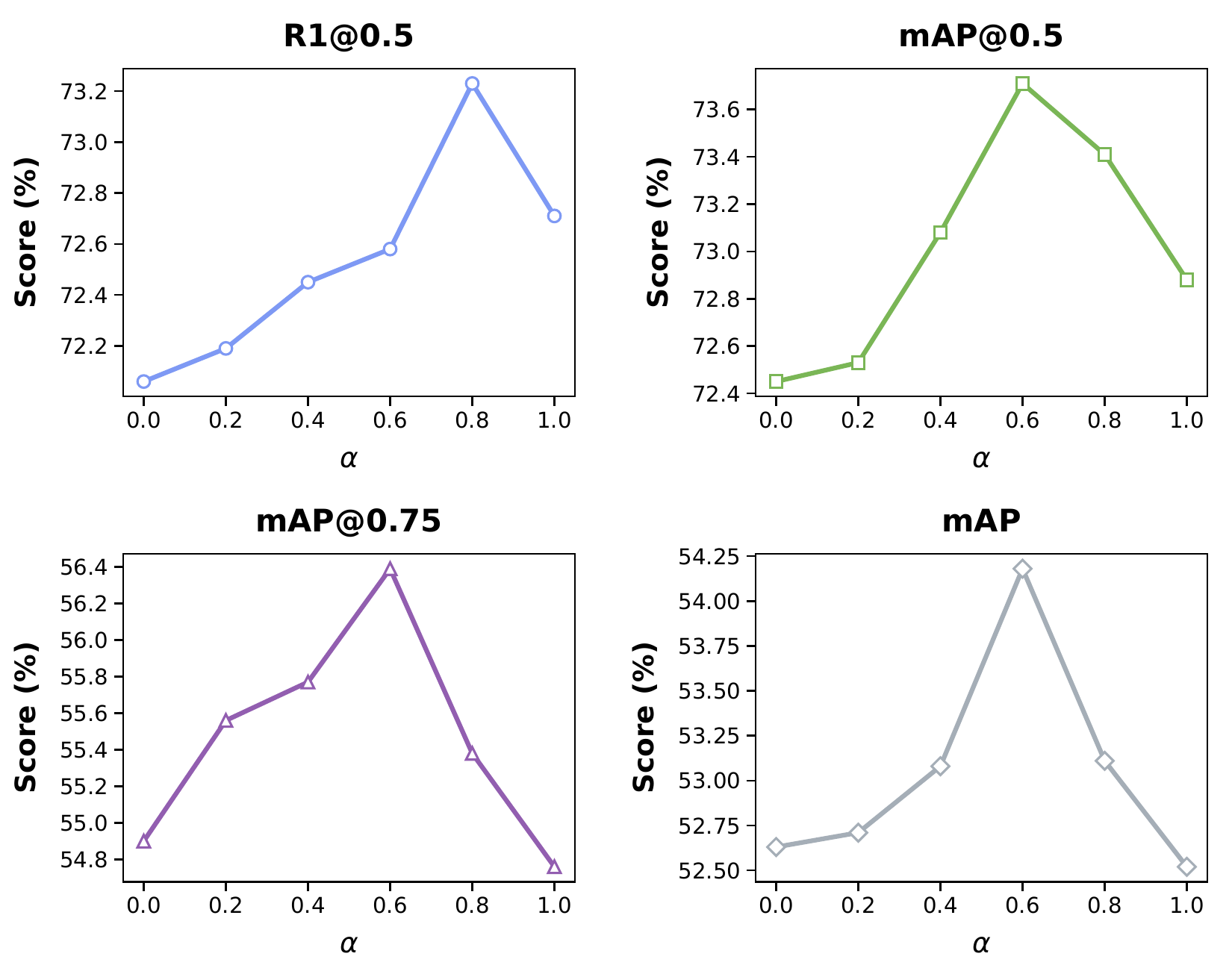}
    \caption{
   Sensitivity analysis of the Interaction weight $\alpha$ on QVHighlights~\cite{Alpher79}. The model maintains stable performance across a wide range of values, demonstrating robustness to hyperparameter selection.
    }
    \label{fig:line}
\end{figure}

\begin{figure}[t]
    \centering
    \includegraphics[width=\linewidth]{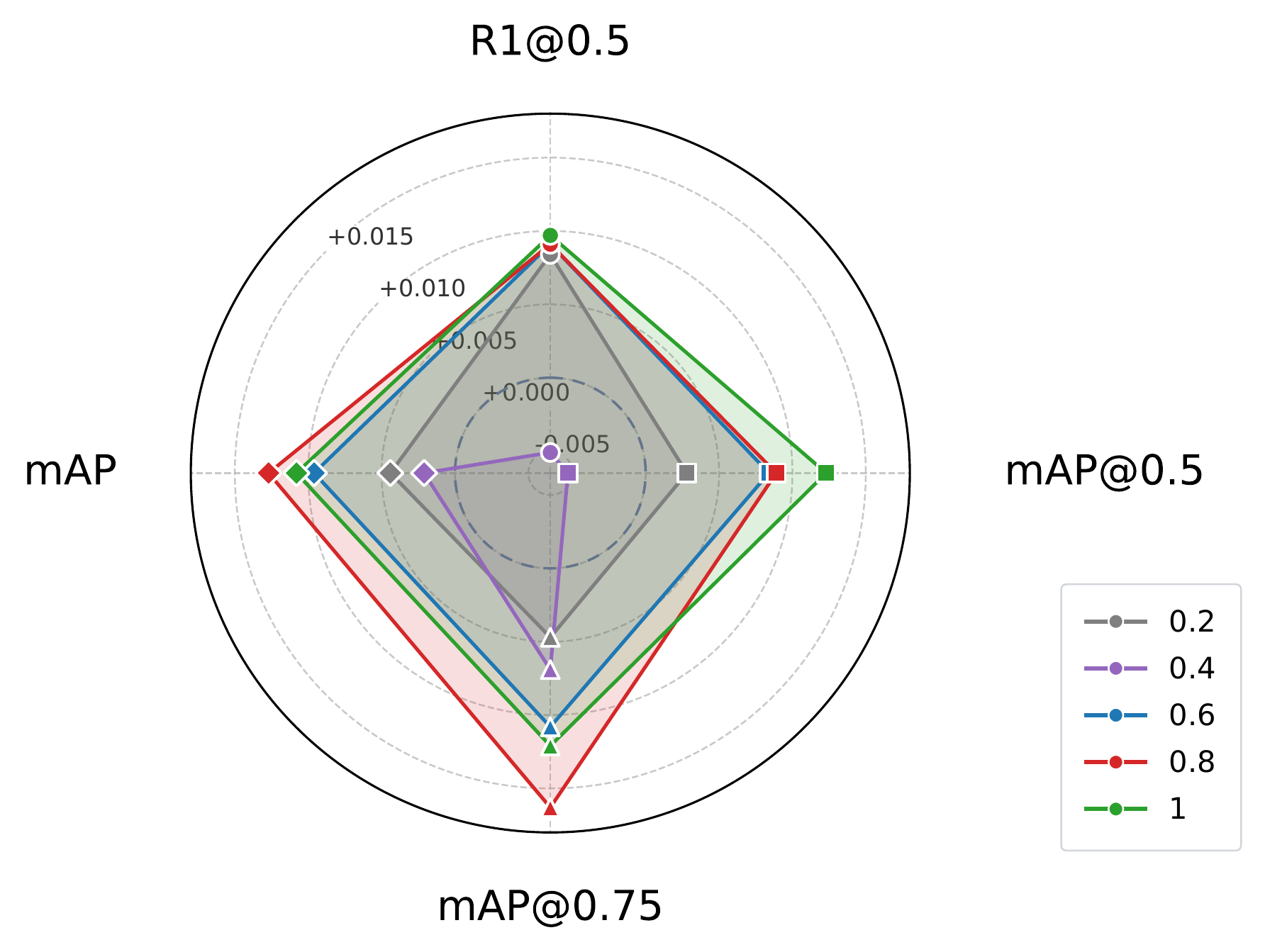}
    \caption{
  Effect of the Interaction-consistency weight $\beta$ 
on QVHighlights~\cite{Alpher79} validation set. We report Recall and mAP.
    }
    \label{fig:radar}
\end{figure}

\subsection{Ablation Study}

\noindent\textbf{Component Ablation.}
As shown in Table~\ref{tab:ablation_main}, progressively enabling FDIM and ISA on QVHighlights yields consistent gains. From a representation perspective, FDIM reduces spurious correlations by explicitly modeling action--entity interactions, enhancing the model's ability to localize correct regions via fine-grained linguistic evidence. From a supervision perspective, ISA further elevates high-precision metrics (e.g., mAP@0.75) by enforcing binding--saliency consistency during training. Notably, ISA improves performance without increasing inference cost, confirming that our gains stem from optimized alignment rather than model capacity.

\noindent\textbf{Robustness to Role-Label Corruption.}
FDIM uses noun and verb masks to construct role-specific interaction
evidence. To test whether the model genuinely depends on correct
linguistic roles rather than merely benefiting from an additional
text pathway, we perturb the role assignments at evaluation time.

For each corruption ratio, we randomly select the corresponding
fraction of noun- or verb-labeled tokens and flip their noun/verb
role assignments. The perturbation modifies only the role masks used
by the interaction branch; the encoded query tokens, the full-sentence
grounding pathway, and all model parameters remain unchanged.
No retraining is performed for the corrupted settings.

\begin{table*}[t]
\centering
\caption{Effect of noun/verb role-label corruption on the
QVHighlights validation split.}
\label{tab:role_corruption}
\footnotesize
\setlength{\tabcolsep}{9pt}
\begin{tabular}{@{}lccccc@{}}
\toprule
\textbf{Corruption ratio}
& \textbf{R1@0.5}
& \textbf{R1@0.7}
& \textbf{mAP@0.5}
& \textbf{mAP@0.75}
& \textbf{Avg. mAP} \\
\midrule
0\% (original roles) & \textbf{73.81} & \textbf{58.52}
& \textbf{73.49} & \textbf{57.16} & \textbf{54.47} \\
20\% & 72.84 & 57.03 & 72.68 & 54.67 & 52.71 \\
50\% & 72.00 & 56.13 & 72.61 & 54.10 & 52.37 \\
100\% & 65.10 & 52.39 & 71.26 & 55.86 & 52.74 \\
\bottomrule
\end{tabular}
\end{table*}

As shown in Table~\ref{tab:role_corruption}, top-1 retrieval
performance deteriorates as role assignments become less reliable,
with the largest reduction under complete corruption. Average mAP
changes less monotonically because the original full-sentence pathway
remains active and can still rank plausible proposals. These results
indicate that accurate role information is particularly important for
selecting the correct interaction-consistent moment, rather than being
the sole source of general video--text relevance.

\noindent\textbf{Effect of Interaction Modeling.}
To verify that gains are not merely from stronger features, we compare IAE-VTG against a baseline augmented with SlowFast motion features (Table~\ref{tab:feature_ablation}). While SlowFast provides richer visual evidence, IAE-VTG consistently outperforms it. This suggests that whereas raw feature enrichment quickly saturates in complex scenes, our structural integration of motion and entity information effectively resolves action--entity binding ambiguities that implicit alignment fails to address.

\noindent\textbf{Robustness to Negative Queries.}
We evaluate IAE-VTG on the NA-VMR task~\cite{Alpher73} to test its ability to reject mismatched queries (Table~\ref{tab:na_vmr}). IAE-VTG achieves substantial gains in Rejection Accuracy (+16.13\% RA-ID, +7.22\% RA-OOD) while simultaneously improving R1 localization. The significant boost in RA-ID improved rejection of semantically plausible mismatched queries.

\noindent\textbf{Sensitivity to Semantic Perturbations.}
We conduct a stress test by perturbing queries through verb,
object, or relation swaps while keeping the video fixed
(Table~\ref{tab:semantic_swap}).
All perturbation types substantially reduce the overlap between
the top-ranked prediction and the original ground-truth moment,
while also decreasing prediction confidence and producing high
rejection rates.
These results indicate that IAE-VTG responds sensitively to
controlled semantic inconsistencies rather than relying solely
on superficial visual cues. 

\noindent\textbf{Hyperparameter Sensitivity.}
The corresponding sensitivity trends of $\alpha$ and $\beta$ are
further analyzed in Figs.~\ref{fig:line} and~\ref{fig:radar}.

\subsection{Generalization and Query Complexity}
\label{sec:generalization_analysis}

\noindent\textbf{Compositional Generalization.}
To examine whether the proposed action--entity binding mechanism
generalizes beyond the standard test distribution, we evaluate the
baseline and IAE-VTG on the \emph{Trivial},
\emph{Novel-Composition} (Novel-C), and \emph{Novel-Word}
(Novel-W) splits introduced by the compositional temporal grounding
protocol of Li \emph{et al.}~\cite{Alpher76}.
These splits evaluate increasingly challenging forms of
compositional generalization, ranging from familiar compositions
to novel combinations and novel lexical elements.

\begin{table}[t]
\centering
\caption{Baseline-to-IAE-VTG comparison on the compositional
temporal grounding splits.}
\label{tab:visa}
\footnotesize
\setlength{\tabcolsep}{4.4pt}
\begin{tabular}{@{}lcccc@{}}
\toprule
& \multicolumn{2}{c}{\textbf{R1@0.7}}
& \multicolumn{2}{c}{\textbf{mIoU}} \\
\cmidrule(lr){2-3}\cmidrule(l){4-5}
\textbf{Split}
& \textbf{Baseline}
& \textbf{IAE-VTG}
& \textbf{Baseline}
& \textbf{IAE-VTG} \\
\midrule
Trivial & 12.16 & \textbf{13.14} & 30.66 & \textbf{31.56} \\
Novel-C & 7.92 & \textbf{8.49} & 24.62 & \textbf{25.17} \\
Novel-W & 8.11 & \textbf{9.46} & 24.73 & \textbf{26.43} \\
\bottomrule
\end{tabular}
\end{table}

As shown in Table~\ref{tab:visa}, IAE-VTG improves both R1@0.7
and mIoU on all three splits. The largest gain occurs on Novel-W,
where generalization requires handling unfamiliar lexical
compositions while preserving the underlying action--entity
structure. These results support compositional transfer without
replacing the original full-sentence representation.

\noindent\textbf{Behavior on Complex Queries.}
We further group queries according to linguistic complexity and
evaluate IAE-VTG on the three sufficiently represented categories
shown in Table~\ref{tab:complex}. A separate negation subset contains
only seven samples ($N=7$), so we do not use it to support a general
category-level conclusion.

\begin{table}[t]
\centering
\caption{Performance on the query-complexity groups.}
\label{tab:complex}
\footnotesize
\setlength{\tabcolsep}{6pt}
\begin{tabular}{@{}lcc@{}}
\toprule
\textbf{Query group} & \textbf{R1@0.7} & \textbf{Avg. mAP} \\
\midrule
Multi-entity & 58.14 & 53.99 \\
Multi-action & 53.87 & 51.79 \\
Complex & 57.79 & 53.60 \\
\bottomrule
\end{tabular}
\end{table}

IAE-VTG maintains comparable performance on multi-entity and
mixed-complexity queries, while multi-action queries remain more
challenging. This behavior is consistent with the intended scope of
the method: the auxiliary role-specific branch improves explicit
action--entity binding but does not replace broader event-level or
paragraph-level reasoning.

\subsection{Extended Comparisons}
\label{sec:extended_comparisons}

\noindent\textbf{Comparison with Large-Model-Based VTG.}
Large multimodal models provide substantially stronger pretraining
and model capacity. Table~\ref{tab:large_models} therefore serves as
a scale-aware comparison rather than a claim of uniform superiority.

\begin{table}[t]
\centering
\caption{Comparison with large-model-based methods on QVHighlights.}
\label{tab:large_models}
\footnotesize
\setlength{\tabcolsep}{4.2pt}
\begin{tabular}{@{}lccc@{}}
\toprule
\textbf{Method} & \textbf{Size} & \textbf{R1@0.5} & \textbf{R1@0.7} \\
\midrule
\multicolumn{4}{c}{\emph{Zero-shot / LLM-based}} \\
TimeSuite~\cite{Alpher84} & 7B & 12.3 & 9.2 \\
UniTime~\cite{Alpher85} & 7B & 41.0 & 31.5 \\
\midrule
\multicolumn{4}{c}{\emph{QVHighlights-trained}} \\
Chrono-BLIP~\cite{Alpher82} & 4B & 76.8 & 62.8 \\
Chrono-Qwen~\cite{Alpher82} & 3B & 79.1 & 64.8 \\
SlotVTG~\cite{Alpher83} & 3B & 79.5 & 64.6 \\
Chrono-Qwen~\cite{Alpher82} & 7B & 81.8 & 67.6 \\
SlotVTG~\cite{Alpher83} & 7B & \textbf{82.9} & \textbf{69.3} \\
IAE-VTG & 13.26M & 73.81 & 58.52 \\
\bottomrule
\end{tabular}
\end{table}

Although the large models obtain higher absolute recall, IAE-VTG
operates with orders-of-magnitude fewer parameters. This comparison
supports a complementary interpretation: large-scale pretraining
provides broad semantic priors, whereas IAE-VTG introduces an
explicit lightweight inductive bias for action--entity consistency.

\noindent\textbf{Comparison with Saliency-Guided DETR Variants.}
SG-DETR~\cite{Alpher86} strengthens temporal grounding through
saliency-guided modules, and its hybrid variant additionally modifies
the detector head.

\begin{table}[t]
\centering
\caption{Comparison with SG-DETR variants on QVHighlights validation.}
\label{tab:sgdetr}
\scriptsize
\setlength{\tabcolsep}{3.0pt}
\begin{tabular}{@{}lccc@{}}
\toprule
\textbf{Metric}
& \textbf{SG-DETR}
& \textbf{SG-DETR + Hybrid}
& \textbf{IAE-VTG} \\
\midrule
R1@0.5 & 72.10 & 72.80 & \textbf{73.81} \\
R1@0.7 & 57.60 & \textbf{59.50} & 58.52 \\
mAP@0.5 & 72.60 & \textbf{73.50} & 73.49 \\
mAP@0.75 & 53.60 & \textbf{57.90} & 57.16 \\
Avg. mAP & 52.20 & \textbf{55.60} & 54.47 \\
\bottomrule
\end{tabular}
\end{table}

IAE-VTG improves over the saliency-guided SG-DETR baseline on all
reported metrics, while the hybrid detector remains stronger overall.
The distinction is informative: the hybrid head primarily improves
proposal coverage and boundary regression, whereas IAE-VTG focuses on
compositional interaction and assignment quality. These directions
are therefore complementary rather than directly interchangeable.

\section{Conclusion}

In this work, we investigated action--entity ambiguity as an
important source of spurious temporal grounding, where a model may
respond strongly to individually relevant actions or entities without
verifying whether they jointly constitute the queried event.
To address this limitation, we proposed IAE-VTG, which models
interaction consistency at both the representation and supervision
levels. FDIM constructs composition-sensitive temporal evidence by
grounding action- and entity-related query information in
complementary motion and appearance streams, while ISA incorporates
the same interaction criterion into bipartite assignment.

Experiments on QVHighlights, Charades-STA, and TACoS demonstrate
competitive or state-of-the-art grounding performance. Component
ablations, role-label corruption, semantic perturbations, and
qualitative analyses further show that the gains arise from explicit
interaction modeling rather than simply increasing feature capacity.
The improvements on compositional splits and complex-query groups
provide additional evidence that the interaction branch generalizes
beyond the standard evaluation distribution while preserving the
original full-sentence pathway.

Overall, the results indicate that explicit action--entity
interaction modeling provides a useful and lightweight complement to
conventional holistic video--text alignment. At the same time,
action--entity ambiguity is not universal to all VTG samples, and
more general graph-based relational reasoning, rare linguistic
phenomena, and paragraph-level dense grounding remain promising
directions for future work.

\bibliographystyle{IEEEtran}
\bibliography{sample-base}



\vfill

\end{document}